\pdfoutput=1

\documentclass[11pt]{article}

\usepackage{EMNLP2022}

\usepackage{times}
\usepackage{latexsym}

\usepackage[T1]{fontenc}

\usepackage[utf8]{inputenc}

\usepackage{microtype}

\usepackage{inconsolata}

\usepackage{graphicx}
\usepackage{amsmath}
\usepackage{multirow}
\usepackage{array}
\usepackage{booktabs}
\usepackage{engord}
\usepackage{bm}
\usepackage{enumitem}
\usepackage{pifont}
\usepackage{amsthm,amsmath,amssymb}
\newcommand{\tabincell}[2]{\begin{tabular}{@{}#1@{}}#2\end{tabular}}
\usepackage{color}
\title{ChatGPT is a Knowledgeable but Inexperienced Solver: An Investigation of Commonsense Problem in Large Language Models}

\author{Ning Bian$^{\rm 1,2}$, Xianpei Han$^{1,2,3,}$\thanks{* Corresponding Author}, Le Sun$^{\rm 1,2,3}$, Hongyu Lin$^{\rm 2}$, Yaojie Lu$^{\rm 2}$, Ben He$^{\rm 1,2}$, \textbf{Shanshan Jiang$^{\rm 4}$, Bin Dong$^{\rm 4}$}\\
  $^{1}$University of Chinese Academy of Sciences, Beijing, China\\
  $^{2}$Chinese Information Processing Laboratory ~ $^{3}$State Key Laboratory of Computer Science \\
  Institute of Software, Chinese Academy of Sciences, Beijing, China \\ 
  $^{4}$Ricoh Software Research Center (Beijing) Co., Ltd., Beijing, China \\
  \texttt{bianning21@mails.ucas.ac.cn \ \{xianpei,sunle,hongyu,yaojie\}@iscas.ac.cn} \\
  \texttt{benhe@ucas.ac.cn \ \{shanshan.jiang, bin.dong\}@cn.ricoh.com} \\}

\begin{document}
\maketitle
\begin{abstract}
Large language models (LLMs) have made significant progress in NLP. However, their ability to memorize, represent, and leverage commonsense knowledge has been a well-known pain point. In this paper, we specifically focus on ChatGPT, a widely used and easily accessible LLM, and ask the following questions: 
(1) Can ChatGPT effectively answer commonsense questions? 
(2) Is ChatGPT aware of the underlying commonsense knowledge for answering a specific question?
(3) Is ChatGPT knowledgeable in commonsense?
(4) Can ChatGPT effectively leverage commonsense for answering questions?
We conduct a series of experiments on 11 datasets to evaluate ChatGPT's commonsense abilities, including answering commonsense questions, identifying necessary knowledge, generating knowledge descriptions, and using knowledge descriptions to answer questions again. 
Experimental results show that:
(1) ChatGPT can achieve good QA accuracies in commonsense tasks, while still struggling with certain domains of datasets.
(2) ChatGPT is knowledgeable, and can accurately generate most of the commonsense knowledge using knowledge prompts.
(3) Despite its knowledge, ChatGPT is an inexperienced commonsense problem solver, which cannot precisely identify the needed commonsense for answering a specific question.
These findings raise the need to explore improved mechanisms for effectively incorporating commonsense into LLMs like ChatGPT, such as better instruction following and commonsense guidance. 
\end{abstract}

\section{Introduction}

Commonsense knowledge is a foundational aspect of human cognition, encompassing our innate understanding of the world and our capacity to reason within it. It includes knowledge about the spatial, physical, social, temporal, and psychological aspects of the typical everyday life, as well as an awareness of social norms, beliefs, and values \cite{liu2004conceptnet, davis2023benchmarks}. The integration of commonsense knowledge is essential for developing NLP systems that can comprehend and generate language similar to humans. However, acquiring and representing commonsense in machines has posed a long-standing challenge \cite{li2021language, zhang2022alleviating, zhou2023commonsense}, primarily due to the implicit and context-dependent nature of commonsense \cite{gordon2013reporting,shwartz2020neural}. In recent years, there has been a growing interest in addressing the commonsense problem within NLP models, with the aim of enabling more human-like language generation and understanding \cite{bauer2018commonsense, wang2020connecting, jiang2021m, liu2021kg, sun2022tsgp, liu2022generated, he2023buca, cui2023free, chen2023distinguish}. 


Recently, large language models (LLMs) such as ChatGPT have made substantial advancements in a wide range of NLP tasks, including inference, contextual understanding \cite{tang2023large}, and chain-of-thought reasoning \cite{wei2022chain}. These achievements suggest that LLMs possess a certain degree of commonsense knowledge \cite{west2022symbolic, gu2022language, zhao2023large}. However, the challenge of commonsense still remains a significant limitation for LLMs \cite{zhou2020evaluating, li2022systematic, bhargava2022commonsense, bang2023multitask, kondo2023probing}. Despite their increasing abilities, the extent of LLMs' understanding and reasoning capabilities regarding commonsense knowledge remains unclear. 

In this paper, we specifically focus on ChatGPT to evaluate the commonsense abilities in LLMs. ChatGPT is a prominent and widely used representative of LLMs, due to its high performance and ease of access. We pose the following key questions on the commonsense abilities of ChatGPT:
(1) \textit{Can ChatGPT effectively answer commonsense questions?}
(2) \textit{Is ChatGPT aware of the underlying commonsense knowledge for answering a specific question?}
(3) \textit{Is ChatGPT knowledgeable in commonsense?}
(4) \textit{Can ChatGPT effectively leverage commonsense for answering questions?}
Answering these questions is crucial for understanding the capabilities and limitations of LLMs and for developing better methods to evaluate and improve their performance on commonsense tasks.
To do so, we employ 11 commonsense QA datasets that cover a wide range of 8 commonsense domains, including physical, social, temporal, and numerical reasoning, etc. Our evaluation methodology consists of four key steps. First, we present commonsense questions to the GPT-3, Instruct GPT (text-davinci-003), and ChatGPT, and assess the accuracy of their responses. This step helps us gauge the models' ability to accurately answer commonsense questions. Next, we investigate whether ChatGPT possesses an understanding of the underlying commonsense knowledge necessary for answering these questions. We prompt the models to describe the required knowledge and evaluate the accuracy and appropriateness of their descriptions. Finally, we explore the models' capacity to leverage commonsense knowledge for reasoning. We utilize the knowledge generated in the previous experiments as context and ask the models to answer the questions again, in order to evaluate whether the models can effectively leverage the identified knowledge in their reasoning process. We further compare their performance using ``golden" knowledge.

Our experiments provide insights into the commonsense problem of ChatGPT: 
(1) ChatGPT and Instruct GPT can achieve good QA accuracies on commonsense tasks, while still struggling with certain domains of datasets.
(2) ChatGPT is knowledgeable and can accurately generate most of the commonsense knowledge using knowledge prompts.
(3)	ChatGPT is an inexperienced commonsense problem solver, which cannot precisely identify the needed commonsense knowledge for solving a specific question. Furthermore, ChatGPT cannot effectively leverage knowledge in context for answering questions.

The main contributions of this paper are:

\begin{itemize}[leftmargin=*]
\parskip -0.4ex
\item We investigate the commonsense ability of ChatGPT in detail by conducting experiments to answer four key questions. 

\item We design a series of experiments to evaluate ChatGPT's ability to memorize, represent and leverage commonsense knowledge, including answering commonsense questions, identifying and generating necessary knowledge, and leveraging commonsense knowledge for reasoning.

\item By identifying the strengths and weaknesses of ChatGPT's ability in commonsense knowledge and reasoning, we provide insights into the development of more advanced language models that can effectively leverage and reason about commonsense knowledge.
\end{itemize}

\section{What is Commonsense?}

\begin{table*}[t]
\centering
\small
\begin{tabular}{ll|m{294pt}}
\hline
Dataset &
  Domain &
  \multicolumn{1}{c}{Example (Bold texts are the answers)} \\ \hline
CommonsenseQA &
  General &
  Choose your answer to the question: Where are you likely to find a hamburger? \textbf{A. fast food restaurant}, B. pizza, C. ground up dead cows, D. mouth, E. cow circus 
   \\ \hline
OpenBookQA &
  General &
  Choose your answer to the question: If a person walks in the opposite direction of a compass arrow they are walking A. west, B. north, C. east, \textbf{D. south}
   \\ \hline
WSC &
  General &
  Choose subsentence A or B that completes the sentence: The trophy doesn't fit into the brown suitcase because A. the trophy is too small. \textbf{B. the suitcase is too small.} 
   \\ \hline
PIQA &
  Physical &
  Choose one that is correct: \textbf{A. ice box will turn into a cooler if you add water to it.} B. ice box will turn into a cooler if you add soda to it. 
   \\ \hline
Social IQA &
  Social &
  Taylor taught math in the schools after studying to be a teacher. Choose the most suitable answer for the question: What does Taylor need to do before this? \textbf{A. get a certificate}, B. teach small children, C. work in a school 
   \\ \hline
ARC &
  Science &
  Choose your answer to the question: Which technology was developed most recently? \textbf{A. cellular telephone}, B. television, C. refrigerator, D. airplane 
   \\ \hline
QASC &
  Science &
  Choose your answer to the question: What is described in terms of temperature and water in the air? A. storms; \textbf{B. climate}; C. mass; D. seasonal; E. winter; F. density; G. length 
   \\ \hline
HellaSWAG &
  Event &
  Choose your answer to the question: We see a chair with a pillow on it. A. a man holding a cat does curling. B. a man holding a cat starts hitting objects on an item. C. a man holding a cat is wrapping a box. \textbf{D. a man holding a cat sits down on the chair.}
   \\ \hline
NumerSense &
  Numerical &
  a square is a shape with \textless mask\textgreater\ equally lengthed sides. (\textbf{four})
   \\ \hline
ProtoQA &
  Prototypical &
  Use simple words separated by commas to name something in your life that could cause you to lose weight. 
  (\textbf{Eating less, exercising more, stress.}) \\ \hline
MC-TACO &
  Temporal &
  Select all feasible answers for the question: Carl Laemmle, head of Universal Studios, gave Einstein a tour of his studio and introduced him to Chaplin. At what time did Einstein return home? \textbf{A. 8:00 PM}; B. a second later; \textbf{C. a hour later}  \\ \hline
\end{tabular}
\caption{\label{t1}
Domains and examples of the commonsense QA datasets used in this paper.
}
\end{table*}

Commonsense knowledge is ``a huge portion of human experience, encompassing knowledge about the spatial, physical, social, temporal, and psychological aspects of typical everyday life'' \cite{liu2004conceptnet,brachman2022new}. This type of knowledge is often taken for granted and is typically acquired through years of experience and socialization within a particular culture. 

To establish the necessary background and preliminary concepts for our study, we summarize several main categories of commonsense as following:
\textbf{General commonsense} refers to knowledge that is widely shared and assumed to be true by most people, such as \textit{the sun rises in the east and sets in the west}.
\textbf{Physical commonsense} involves intuitive knowledge about the physical world, such as \textit{objects fall to the ground when dropped} and \textit{water flows downhill}.
\textbf{Social commonsense} involves knowledge about social norms, customs, and practices, such as \textit{it is polite to say ``thank you'' when making requests}.
\textbf{Science commonsense} involves knowledge about basic scientific principles, such as \textit{gravity pulls all objects on Earth to Earth's center}.
\textbf{Event commonsense} involves knowledge about the sequence of events and the causal relationships between them, such as \textit{if a glass is knocked over, the liquid inside will spill}.
\textbf{Numerical commonsense} involves knowledge about numbers, such as \textit{human has two hands and ten fingers}.
\textbf{Prototypical commonsense} involves knowledge about typical or prototypical examples of concepts, such as \textit{a swallow is a kind of bird, and a bird has wings}.
\textbf{Temporal commonsense} involves knowledge about time, such as \textit{traveling abroad requires a longer time than taking a walk}.

\section{Can ChatGPT Effectively Answer Commonsense Questions?}\label{s3}

In this section, we evaluate the performance of ChatGPT to answer commonsense questions. We use 11 commonsense QA datasets covering 8 commonsense domains, including general, physical, social, science, event, numerical, prototypical, and temporal knowledge. The 11 datasets are CommonsenseQA \cite{talmor2019commonsenseqa}, OpenBookQA \cite{mihaylov2018can}, WSC \cite{levesque2012winograd}, PIQA \cite{bisk2020piqa}, Social IQA \cite{sap2019social}, ARC (Easy set) \cite{clark2018think}, QASC \cite{khot2020qasc}, HellaSWAG \cite{zellers2019hellaswag}, NumerSense \cite{lin2020birds}, ProtoQA \cite{boratko2020protoqa}, and MC-TACO \cite{zhou2019going}. The datasets, domains, and an example for each dataset are shown in Table \ref{t1}. 

We sample 100 questions from the development set of each dataset, except for ProtoQA, which has only 52 questions in its development set. We use GPT-3 (davinci, \citealp{brown2020language}), Instruct GPT (text-davinci-003), and ChatGPT (we use the ``GPT-3.5'' web interface\footnote{chat.openai.com}) for evaluation. For GPT-3, we use 4-shot in-context learning, as GPT-3 cannot effectively answer questions in a zero-shot setting. For Instruct GPT and ChatGPT, we use zero-shot inference and design prompt templates for different datasets (shown in Table \ref{t1}).

From results in Table \ref{t2}, we can see that:

\begin{table*}[!t]
\centering
\small
\begin{tabular}{llcccc}
\hline
Dataset       & Domain       & GPT-3   & Instruct GPT       & ChatGPT         & Human       \\ \hline
CommonsenseQA & General      & 38.0    & \textbf{81.0} & 74.0            & 88.9        \\
OpenBookQA    & General      & 22.0    & 65.0          & \textbf{73.0}   & 89.3        \\
WSC           & General      & 46.0    & \textbf{78.0} & \textbf{78.0}   & 92.1        \\
PIQA          & Physical     & 48.0    & 77.0          & \textbf{78.0}   & 94.5        \\
Social IQA    & Social       & 36.0    & \textbf{71.0} & 62.0            & 86.9        \\
ARC           & Science      & 27.0    & 88.0          & \textbf{94.0}   & --          \\
QASC          & Science      & 25.0    & \textbf{75.0} & 74.0            & 93.0        \\
HellaSWAG     & Event        & 19.0    & 61.0          & \textbf{67.0}   & 95.7        \\
NumerSense    & Numerical    & 45.0    & 63.0          & \textbf{79.0}   & 89.7        \\
ProtoQA       & Prototypical & 67.3    & 84.6          & \textbf{94.2}   & --          \\
MC-TACO       & Temporal     & 20.0    & \textbf{53.0} & 52.0            & 75.8        \\ \hline 
\end{tabular}
\caption{\label{t2}
Evaluation results (\% accuracy) of large language models on commonsense QA datasets. Human accuracies are adopted from the original papers of these datasets except WSC by \citet{bender2015establishing}.
}
\end{table*}

\textbf{Instruct GPT and ChatGPT demonstrate high accuracy in answering commonsense questions.} We evaluate the performances of different LLMs on 11 commonsense QA datasets. The results in Table \ref{t2} show that both Instruct GPT and ChatGPT achieve good performance across most datasets. 
Notably, ChatGPT achieves the highest accuracy of 94\% on the ARC dataset and 94.2\% on the ProtoQA dataset. This indicates that ChatGPT is capable of accurately answering various types of commonsense questions. However, there are still large accuracy gaps between models and humans, as shown in Table \ref{t2}.

\textbf{The ability of models to leverage commonsense is probably improved by instruction tuning and human alignment.} A notable observation from the results in Table \ref{t2} is the significant improvement achieved by Instruct GPT and ChatGPT compared to GPT-3. This improvement is probably due to the incorporation of instruction tuning and human alignment during training \cite{ouyang2022training}. In addition to the pre-training, these techniques may enable the models to better leverage and reason with commonsense knowledge, demonstrating the importance of instruction and alignment in enhancing the models' performance. 


Overall, ChatGPT achieves higher accuracy than Instruct GPT in most domains, demonstrating the effectiveness of the RLHF technique in enhancing knowledge-leveraging ability. However, Instruct GPT slightly outperforms ChatGPT on certain datasets including CommonsenseQA (p = 0.238 by T-test) and Social IQA (p = 0.179). This is because ChatGPT tends to be cautious and refuses to answer questions when information is insufficient, resulting in outputs like ``Based on the information given, it is not possible to determine ...''. These outputs are considered incorrect because they indicate an inability to generate a direct and correct answer. In CommonsenseQA, ChatGPT produces 4 such responses out of 100 questions, while in Social IQA, there are 13 such instances out of 100. This highlights the need for further research on how to balance the caution and accuracy in models where there is insufficient information. Achieving this requires the model to understand the necessary knowledge to answer the question and be aware of whether the model possesses that knowledge.

\textbf{Although ChatGPT performs well on most commonsense knowledge domains, they still struggle with certain domains.} The experiments in Table \ref{t2} show that ChatGPT lags behind on datasets regarding social, event, and temporal (Social IQA, HellaSWAG, and MC-TACO datasets), with the ChatGPT's performances below 70\%. This shows that ChatGPT still has drawbacks on the social, event, and temporal domains of commonsense QA, which is consistent with \citet{chang2023language}. We believe this is because these domains of commonsense require a deeper understanding of human behavior and social interactions, and they appear infrequently in text corpora. ChatGPT needs to go beyond superficial semantic understanding and learn about human behaviors to better learn these domains of commonsense.

\section{Is ChatGPT Aware of the Commonsense needed for QA?}\label{s4}

\begin{table}[!t]
\centering
\small
\begin{tabular}{|m{200pt}|}
\hline
\multicolumn{1}{|l|}{
    \tabincell{m{200pt}}{ 
    \specialrule{0em}{0pt}{3pt}
    \textbf{Question}: How to make outdoor pillow? \\
 A. Blow into \emph{tin can} and tie with rubber band. \ding{55} \\
 B. Blow into \emph{trash bag} and tie with rubber band. \ding{51}
    }} \\
\multicolumn{1}{|l|}{
    \tabincell{m{200pt}}{ 
    \specialrule{0em}{0pt}{3pt}
    \textbf{ChatGPT Response}: \\
			1. Understanding of what an outdoor pillow is. \ding{51}\\
			2. Knowledge of how to make an outdoor pillow.\\
			3. Familiarity with the materials required to make an outdoor pillow. \ding{51}\\
			4. Understanding of how to inflate a container such as a tin can or trash bag.\\
			5. Knowledge of how to tie a rubber band.\\
    }} \\
\multicolumn{1}{|l|}{
    \tabincell{m{200pt}}{ 
    \specialrule{0em}{0pt}{3pt}
    \textbf{Human Evaluation}: \\
			1. Understanding of what an outdoor pillow is. \\
			3. Familiarity with the materials required to make an outdoor pillow. \\
			a. Knowing that tin can is hard and trash bag is soft.\\
			b. Knowing that human prefers soft pillows.
    }} \\
\hline
\end{tabular}
\caption{\label{t3}
An example of necessary knowledge generated by ChatGPT and human evaluation. The question is from the PIQA dataset.
}
\end{table}
 
\begin{table*}[!t]
\centering
\small
\begin{tabular}{llccc}
\hline
Dataset       & Domain       & Correct              & Wrong                 & Overall\\ \hline
CommonsenseQA & General      & 65.83 / 94.17 / 75.86    & 50.00 / 72.50 / 57.79     & 57.92 / 83.33 / 66.82\\
OpenBookQA    & General      & 80.50 / 100.00 / 87.94   & 35.83 / 55.83 / 42.81     & 58.17 / 77.92 / 65.37\\
WSC           & General      & 80.00 / 87.50 / 83.21    & 57.50 / 83.33 / 65.90     & 68.75 / 85.12 / 74.56\\
PIQA          & Physical     & 60.00 / 80.00 / 64.90    & 53.36 / 88.33 / 63.25     & 56.78 / 84.17 / 64.08\\ 
Social IQA    & Social       & 53.00 / 90.00 / 63.43    & 28.17 / 40.00 / 32.05     & 40.58 / 65.00 / 47.74\\
ARC           & Science      & 73.57 / 100.00 / 82.80   & 45.00 / 83.33 / 55.36     & 65.00 / 95.00 / 74.57\\
QASC          & Science      & 67.17 / 100.00 / 78.79   & 68.33 / 88.33 / 73.48     & 67.75 / 94.17 / 76.13\\
HellaSWAG     & Event        & 64.00 / 95.00 / 74.10    & 47.55 / 73.00 / 57.31     & 55.77 / 84.00 / 65.70\\
NumerSense    & Numerical    & 44.00 / 95.00 / 58.29    & 44.00 / 89.17 / 58.21     & 44.00 / 92.08 / 58.25\\
ProtoQA       & Prototypical & 65.88 / 98.04 / 76.96    & 48.33 / 88.89 / 58.73     & 63.25 / 96.67 / 74.23\\
MC-TACO       & Temporal     & 47.50 / 80.00 / 58.00    & 26.17 / 61.67 / 35.57     & 36.83 / 70.83 / 46.79\\ \hline
\end{tabular}
\caption{\label{t4}
Precision / Recall / F1 scores of ChatGPT generated necessary knowledge for correct- and wrong-answered questions.
}
\end{table*}

In Section \ref{s3}, we found that ChatGPT performs well on commonsense QA datasets. This intrigues us to explore whether ChatGPT is experienced experts that are aware of what knowledge is needed and can leverage the knowledge for question answering, or if they are inexperienced problem solvers that rely on memorizing a large amount of information that covers the answers. 

To answer this question, we sample 20 questions from each commonsense QA dataset and ask ChatGPT ``What knowledge is necessary for answering this question? \{question\} \{answer choices (if applicable)\}''. We sample 10 correctly and 10 incorrectly answered questions for each dataset to ensure a fair comparison.
In cases where there are insufficient incorrectly answered questions (specifically, there are 6 for ARC and 3 for ProtoQA), we take all incorrectly answered questions and sample more correctly answered questions to fill up the 20 questions. In total, ChatGPT identified 855 pieces of knowledge as necessary for answering these questions, with an average of 3.9 pieces of knowledge per question.

Human annotators with a solid understanding of commonsense reasoning are employed to manually evaluate the precision and recall of each generated piece of knowledge. Precision refers to the proportion of relevant knowledge correctly included in the response, while recall refers to the extent to which the necessary knowledge is appropriately covered. To synthesize the precision and recall scores into a single metric, we calculate the F1 score for each response. The annotators are provided with the question, the model's response, and the corresponding answer choices (if applicable). They are guided by predefined criteria for evaluating the responses. Specifically, they first assess whether each piece of knowledge is necessary for answering the question. Then, they judge whether the question is answerable based on reasoning upon the labeled necessary knowledge. If the question is still unanswerable, the annotators need to fill in the missing knowledge to answer the question.

For example, Table \ref{t3} shows a response of ChatGPT that outlines the necessary knowledge for answering a specific question. During the manual evaluation, expert annotators assess the usefulness of each piece of knowledge. In this example, knowledge pieces 1 and 3 are labeled as relevant and useful for answering the question accurately. Knowledge 2 is identified as overgeneralized, as it merely repeats information already presented in the question without providing additional insights and details. Knowledge 4 and 5 are labeled as unnecessary for answering the question, as they cannot contribute to distinguishing between answer options. After this assessment, the annotators determine whether the question can be answered based on the two identified pieces of useful knowledge (1 \& 3). However, it is found that additional knowledge is still needed to provide a correct answer. Consequently, two additional pieces of knowledge, a and b, are added to supplement the reasoning process. Thus, the precision and recall for this response are 2/5 and 2/4, so the F1 score is 44.44\%.
The results are shown in Table \ref{t4}. We can see that:

\textbf{ChatGPT is an inexperienced problem solver, which struggles to precisely identify the necessary commonsense knowledge to answer a specific question.} As shown in Table \ref{t4}, the overall F1 scores range from 45\% to about 75\% on most of the commonsense QA datasets. These results demonstrate that ChatGPT is an inexperienced problem solver and cannot accurately identify the necessary knowledge to answer a specific commonsense question. 

Specifically, the model performs relatively well in the science domain, achieving 74.57\% F1 on ARC and 76.13\% on QASC. However, the model exhibits the lowest performances on social and temporal domains, i.e., Social IQA and MC-TACO, which is consistent with the results in Section \ref{s3}. This discrepancy in F1 scores is likely because scientific commonsense knowledge is more prevalent in the text corpus than social and temporal knowledge. For instance, textbooks frequently discuss scientific concepts such as ``\textit{climate is described by temperature and humidity}'', but rarely mention social commonsense like ``\textit{students don't like taking big exams}''. This shows a deficiency of LLMs like ChatGPT in social and temporal knowledge, highlighting the importance of developing more effective training strategies to inject these domains of commonsense knowledge.

\textbf{ChatGPT cannot effectively distinguish between relevant and irrelevant commonsense knowledge for a specific question, which usually generates knowledge with noise.} As shown in Table \ref{t4}, the precisions are significantly higher than the recalls of the generated necessary knowledge (p < 0.05 on all datasets except OpenBookQA with p = 0.09). The average recall of all datasets is 84.42\% and the average precision is 55.88\%. This shows that while the model can identify most of the commonsense knowledge for questions, it struggles with accurately identifying which pieces of knowledge are essential and usually generates irrelevant or overgeneralized knowledge. For example, knowledge 2 in Table \ref{t3} is overgeneralized because the question itself is ``how to make outdoor pillow'', and knowledge 4 and 5 are irrelevant given the difference between the two choices. We believe this is because the model relies on keyword and topic matching, rather than a full understanding of the logical relationships within the question. This further demonstrates that ChatGPT is still inexperienced problem solvers. Therefore, LLMs need to increase their self-awareness of the task at hand and distinguish crucial information from irrelevant background information. They should determine which parts of a question are necessary for decision-making, such as ``blow into tin can/trash bag'' in Table \ref{t3}, and which parts are extraneous.

\begin{figure}[!t]
  \centering
  \includegraphics[width=\columnwidth]{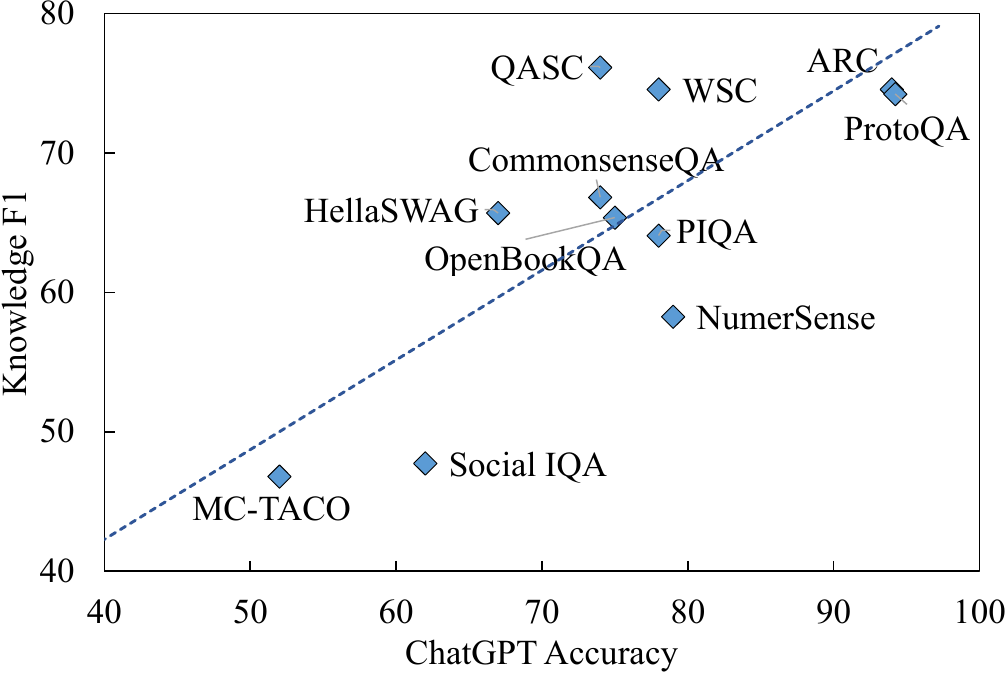}
  \caption{Correlation between necessary knowledge F1 and the answer accuracy of ChatGPT.}
  \label{figure-f1}
\end{figure}

\textbf{The commonsense ability of ChatGPT can be improved by enhancing their knowledge-awareness.} Figure \ref{figure-f1} demonstrates a strong correlation between the overall F1 scores of generated necessary knowledge and the answer accuracies of ChatGPT, with a Pearson coefficient of 0.77 (p = 0.006). Furthermore, Table \ref{t4} shows that the knowledge F1 scores for correctly answered questions are significantly higher than those for incorrectly answered questions (p < 0.05 on OpenBookQA, WSC, Social IQA, ARC, and MC-TACO datasets). These findings suggest that accurately identifying necessary knowledge is crucial for correctly answering commonsense questions. Consequently, enhancing the model's self-awareness of necessary knowledge may improve its performance on downstream tasks including commonsense QA.

\section{Is ChatGPT Knowledgeable in Commonsense?}\label{s5}

\begin{table}[!t]
\centering
\small
\begin{tabular}{|m{200pt}|}
\hline

\multicolumn{1}{|l|}{
    \tabincell{m{200pt}}{ 
    \specialrule{0em}{0pt}{3pt}
    \textbf{Question 1}: Do you have an understanding of what an outdoor pillow is? \\
	\textbf{ChatGPT Response}: \\
\ \ As an AI language model, I have an understanding that an outdoor pillow is a type of pillow that is designed to be used outside, such as on a patio or deck, and is typically made of weather-resistant materials to withstand exposure to the elements.

\ \ Regarding the two options you provided, neither option appears to be a conventional method for making an outdoor pillow. In option A, blowing into a tin can and tying it with a rubber band does not create a cushion-like surface suitable for use as a pillow. In option B, blowing into a trash bag and tying it with a rubber band may create a cushion-like surface, but it is unlikely to be durable or comfortable for prolonged use as an outdoor pillow.

\ \ Therefore, I would suggest exploring other methods or materials that are specifically designed for making outdoor pillows, such as using weather-resistant fabrics and stuffing materials.
    }} \\

\multicolumn{1}{|l|}{
    \tabincell{m{200pt}}{ 
    \specialrule{0em}{0pt}{3pt}
    ......
    }} \\

%
%
%
\hline
\end{tabular}
\caption{\label{t5}
An example of generated knowledge by ChatGPT. The question is in Table 3.
}
\end{table}

This section answers the question: \textit{To what extent do ChatGPT possess commonsense knowledge?} To answer this question, similar to \citet{shwartz2020unsupervised}, we manually construct knowledge-querying prompts based on the generated necessary knowledge in Section \ref{s4}. For example, as shown in Table \ref{t5}, based on knowledge 1 in Table \ref{t3}, we ask ChatGPT knowledge-querying questions like ``Do you have an understanding of what an outdoor pillow is?'' and manually label each generated knowledge description as correct or incorrect. We collect a total of 775 knowledge descriptions for the questions used in the experiments, with an average of 3.5 descriptions per question. The average length of the knowledge descriptions is 136.1 words.

Table \ref{t5} shows an example of a knowledge-querying question and the generated knowledge description. The description says ``\textit{blowing into a trash bag and tying it with a rubber band may create a cushion-like surface, but it is unlikely to be durable or comfortable for prolonged use as an outdoor pillow}'', but it contradicts with the correct answer. So, this description is labeled as incorrect.

From the results in Table \ref{t6} we can see that:

\textbf{ChatGPT is knowledgeable and contains most of the commonsense knowledge for accurately answering questions.} The results in Table \ref{t6} show that the generated knowledge descriptions of ChatGPT can achieve over 70\% accuracy on most commonsense QA datasets, achieving an average accuracy of 82.66\%. This means ChatGPT can generate accurate commonsense knowledge descriptions given knowledge-querying questions. Thus, ChatGPT can serve as commonsense knowledge bases and provide support for downstream tasks. However, the accuracy is low on Social IQA (54.92\%). We believe this is because social commonsense, such as ``\textit{The person who receives help, rather than gives it, should say thank you}'', is not commonly described in texts. This highlights the importance of developing specific approaches to inject social commonsense knowledge into LLMs.

\begin{table}[!t]
\centering
\small
\begin{tabular}{lccc}
\hline
Dataset       & Correct & Wrong     & Overall\\ \hline
CommonsenseQA & 100.00   & 83.83     & 91.92\\
OpenBookQA    & 84.83   & 100.00     & 92.42\\
WSC           & 90.00   & 74.17     & 82.08\\
PIQA          & 85.00   & 62.14     & 73.57\\
Social IQA    & 58.33   & 51.50     & 54.92\\
ARC           & 91.67   & 97.62     & 95.83\\
QASC          & 88.33   & 89.17     & 88.75\\
HellaSWAG     & 80.00   & 70.83     & 75.42\\
NumerSense    & 85.17   & 84.50     & 84.83\\
ProtoQA       & 80.29   & 100.00     & 83.25\\
MC-TACO       & 95.00   & 77.50     & 86.25\\ \hline
\end{tabular}
\caption{\label{t6}
Accuracies (\%) of ChatGPT generated knowledge descriptions for correct- and wrong-answered questions.
}
\end{table}

\begin{figure}[t!]
  \centering
  \includegraphics[width=\columnwidth]{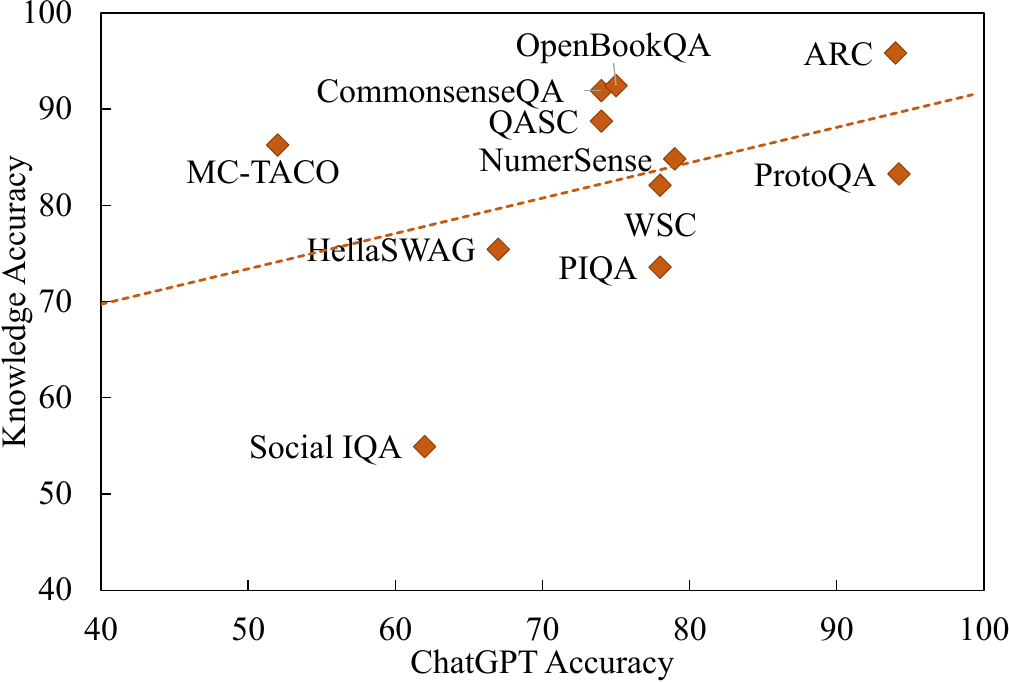}
  \caption{Correlation between generated knowledge accuracy and answer accuracy of ChatGPT.}
  \label{figure-f2}
\end{figure}

\textbf{ChatGPT contains misleading and overgeneralized commonsense knowledge.} We further conduct a manual evaluation of the relevance and informativeness of the knowledge descriptions. We find that 26.25\% of the descriptions include irrelevant and misleading information, and 15.00\% of the descriptions are overgeneralized and fail to provide the specific knowledge necessary to answer the question. Overgeneralized knowledge means correct but unhelpful or irrelevant general knowledge for the given questions. For example, the description in Table \ref{t5} mentions ``exploring other methods or materials that are specifically designed for making outdoor pillows'', which is unhelpful and misleading for answering the question. We believe this is because of noisy and redundant information in the training data of LLMs, which impairs the ability to accurately judge the relevance of information. These findings emphasize the need for instructing LLMs to generate relevant and informative knowledge descriptions that are helpful for QA.

\textbf{There is a gap between knowing and leveraging commonsense knowledge in ChatGPT.} We conduct a correlation analysis between the accuracy of generated knowledge descriptions and answer accuracy, shown in Figure \ref{figure-f2}. Our result reveals a weak correlation between the two accuracies, with a Pearson coefficient of 0.40 (p = 0.227). Notably, while the answer accuracies are low on both the Social IQA and the MC-TACO datasets, there was a significant difference in the accuracy of knowledge descriptions between them: it was low for Social IQA (54.92\%) but high for MC-TACO (86.25\%). Table \ref{t6} further shows that the difference in description accuracy between correctly and incorrectly answered questions is relatively small compared to the results in Table \ref{t4} (p > 0.05 for all datasets except CommonsenseQA with p = 0.01). This shows that a good knowledge description does not necessarily translate to a correct answer. We believe this is because answering commonsense questions not only requires knowledge but also other abilities like reasoning and making inferences under insufficient information conditions.

\section{Can ChatGPT Effectively Leverage Commonsense for Reasoning?}

This section answers the question: \textit{Can ChatGPT leverage commonsense knowledge in context for reasoning and answering questions?} After answering the knowledge-querying questions in Section \ref{s5}, we ask the model to answer the commonsense questions again given the generated knowledge descriptions as context, and evaluate whether the answers will change. 
Specifically, we added these knowledge descriptions before the prompts used in Section \ref{s3}. The prompts remained the same, and the knowledge descriptions served as additional context to evaluate the impact of knowledge on answer changes. This minimal interference with the prompts aims to facilitate a fair comparison between responses with and without contextual knowledge.

Results in Table \ref{t8} show that:

\begin{table}[!t]
\centering
\small
\begin{tabular}{|m{200pt}|}
\hline
\multicolumn{1}{|l|}{
    \tabincell{m{200pt}}{ 
    \specialrule{0em}{0pt}{3pt}
    \textbf{Question}: How to make outdoor pillow? \\
 A. Blow into \emph{tin can} and tie with rubber band. \ding{55} \\
 B. Blow into \emph{trash bag} and tie with rubber band. \ding{51}
    }} \\
\multicolumn{1}{|l|}{
    \tabincell{m{200pt}}{ 
    \specialrule{0em}{0pt}{3pt}
	\textbf{ChatGPT's Answer before Knowledge Description}: \\
Neither option is a good way to make an outdoor pillow. Blowing into a tin can or a trash bag and tying it with a rubber band will not provide adequate support or comfort for an outdoor pillow. Over time, the air will leak out and the pillow will become flat and uncomfortable. A better option for making an outdoor pillow would be to use durable and weather-resistant materials such as outdoor fabric and stuffing or foam designed for outdoor use. Sewing or using fabric glue to attach the fabric together and stuffing it with the appropriate material will result in a more functional and comfortable outdoor pillow.
    }} \\
\multicolumn{1}{|l|}{
    \tabincell{m{200pt}}{ 
    \specialrule{0em}{0pt}{3pt}
	\textbf{ChatGPT's Answer after Knowledge Description}: \\
Neither of these options is a suitable way to make an outdoor pillow. Inflating a tin can or a trash bag and tying it with a rubber band would not provide the necessary level of firmness or support for a pillow, and the materials used may not be durable enough to withstand exposure to outdoor elements.

A better option for making an outdoor pillow would be to use weather-resistant fabric and outdoor-rated stuffing or foam inserts, as I mentioned earlier. This will ensure that the pillow is comfortable, durable, and able to withstand exposure to the elements.
    }} \\

\hline
\end{tabular}
\caption{\label{t7}
An example of the answer before and after knowledge descriptions by ChatGPT. The question is from the PIQA dataset.
}
\end{table}

\textbf{ChatGPT cannot effectively leverage the generated commonsense descriptions if we only add them to the question context.} Our analysis of answer changes before and after using knowledge descriptions shows that in most datasets there are no obvious and consistent accuracy improvements given the commonsense descriptions in contexts. Table \ref{t7} shows an example that a previously incorrect answer remains unchanged after generating knowledge descriptions. There are both wrong-to-correct changes and correct-to-wrong changes, as well as a significant proportion of unchanged answers, for the example shown in Table \ref{t7}. In the case of the Social IQA dataset, the accuracy of knowledge generation is low, leading to more correct answers being modified to become wrong. This shows that ChatGPT cannot effectively exploit its own generated knowledge descriptions to answer questions, and the accuracy of the knowledge generation has a huge impact on the answering results. We believe this is because the model already possesses the generated knowledge, thus adding redundant knowledge is not useful.

\textbf{ChatGPT's performance improvement in commonsense QA is not significant even using golden knowledge.} We use two human-annotated commonsense explanation datasets for the CommonsenseQA dataset, CoS-E \cite{rajani2019explain} and ECQA \cite{aggarwal2021explanations}, as the golden knowledge in context and ask the ChatGPT to generate the answers. We discover that there are only 4/10 wrong $\to$ correct answers given CoS-E explanations, and 8/10 wrong $\to$ correct answers given ECQA explanations while with 1/10 correct $\to$ wrong answer. This shows that ChatGPT cannot answer all questions correctly even given the golden knowledge explanations. We believe this is because ChatGPT lacks the ability to use knowledge for complex commonsense reasoning, such as negation. For example, here is a question that requires reasoning of negation: ``\textit{What would not be true about a basketball if it had a hole in it but it did not lose its general shape? A. punctured, B. popular in America, \textbf{C. full of air}, D. gone, E. round}''. The CoS-E explanation for this question is ``\textit{Air cannot stay in any object that has a hole in it.}'', but ChatGPT still predicts the wrong answer A and explains ``\textit{If a basketball has a hole in it, it is punctured and air can escape from it.}''. These results suggest that LLMs require further guidance and improvement to better leverage and reason about commonsense knowledge in context.

\begin{table}[!t]
\centering
\small
\begin{tabular}{lcc}
\hline
Dataset       & C $\to$ W &W $\to$ C  \\ \hline
CommonsenseQA & 1/10   & 2/10     \\
OpenBookQA    & 0/10   & 5/10     \\
WSC           & 2/10   & 1/10     \\
PIQA          & 1/10   & 1/10     \\
Social IQA    & 5/10   & 1/10     \\
ARC           & 0/14   & 2/6     \\
QASC          & 1/10   & 4/10     \\
HellaSWAG     & 3/10   & 4/10     \\
NumerSense    & 4/10   & 4/10     \\
ProtoQA       & 3/17   & 2/3    \\
MC-TACO       & 4/10   & 2/10     \\ \hline
\end{tabular}
\caption{\label{t8}
Numbers of changed/total answers after the generation of knowledge descriptions. C $\to$ W means a correct answer changes to a wrong answer, and W $\to$ C means a wrong answer changes to a correct answer.
}
\end{table}

\section{Related Work}

Recent studies have shown that LLMs such as GPT-3, ChatGPT, and GPT-4 have made significant progress in various NLP tasks, including QA, text generation, and translation \cite{brown2020language}. However, there is a growing concern about their ability to understand and reason about commonsense knowledge \cite{tamborrino2020pre,cui2021commonsense,bhargava2022commonsense}. Recent studies have focused on evaluating the ability of LLMs to understand commonsense knowledge \cite{davison2019commonsense, liu2020commonsense, niu2021semantic, ma2021exploring, klein2021towards, porada2021does, laskar2023systematic}. For example, \citet{zhou2020evaluating} evaluates several LLMs on a set of commonsense reasoning tasks and found that they have a certain degree of commonsense knowledge, but there is still a gap between models and humans. \citet{wang2021language} studies the generalizability of models for commonsense inference and found that the ability relies heavily on whether the objects to predict are seen during training. \citet{cohn2023dialectical} conduct qualitative investigations of the spatial commonsense reasoning ability of ChatGPT and Bard.
In this paper, we evaluate the commonsense abilities of ChatGPT including answering commonsense questions, identifying and generating necessary knowledge, and leveraging knowledge for reasoning.

\section{Conclusions and Discussions}

In this paper, we investigate the commonsense abilities of ChatGPT and found that ChatGPT is a knowledgeable but inexperienced problem solver: (1) While ChatGPT can achieve good accuracies in commonsense QA, it still struggles with certain domains of QA, including social and temporal commonsense. (2) ChatGPT is knowledgeable in commonsense, which can accurately generate most of the commonsense knowledge using knowledge prompts. (3) ChatGPT is an inexperienced commonsense problem solver. It struggles to precisely identify the underlying commonsense knowledge for a given question and often generates knowledge with a high noise rate. Furthermore, ChatGPT cannot effectively leverage commonsense knowledge in contexts to answer commonsense questions.

The above findings raise several promising directions for the future of LLMs:

(1) Although current ChatGPT is knowledgeable, they are still not experienced problem solvers. Therefore, it is critical to investigate better mechanisms for utilizing commonsense knowledge in LLMs, such as instruction tuning, better commonsense-guided reasoning, etc.

(2) There are still several types of commonsense knowledge missing in LLMs, such as social and temporal commonsense. Therefore it is critical to design knowledge injection approaches for these knowledge types. Furthermore, it is important to design lightweight commonsense updating methods to keep the knowledge up-to-date.

(3) Because ChatGPT does not release its full details, such as training data, hyper-parameters, and checkpoints, and evaluating an ``artificial general intelligence'' model is very difficult, it is crucial to construct benchmarks with wider coverage, and design evaluation methods that provide a more comprehensive and unbiased assessment of LLMs.

\section*{Limitations}

This study provides valuable insights into the commonsense abilities of ChatGPT, but there are several limitations that could be acknowledged and addressed in future research. 

Human evaluations of LLMs' commonsense performance and abilities, such as answer accuracy and the F1 score and accuracy of generated necessary knowledge, are labor-intensive and time-consuming. The manual analysis in this paper required approximately 80 to 100 human-hours in total. Additionally, it can be difficult even for humans to clarify which pieces of commonsense knowledge are necessary for answering a specific question, as commonsense knowledge is often implicit and automatic for humans \cite{ellis2008implicit}. Future studies should explore automated methods for evaluating LLMs' performance and assessing their commonsense abilities. For example, researchers could develop methods that leverage knowledge retrieval or knowledge graphs to evaluate the generated knowledge of LLMs.

Our evaluations of ChatGPT use a small number of sampled commonsense questions on each dataset. While this approach allows for a comprehensive analysis of the commonsense abilities of ChatGPT on different domains of commonsense questions without overwhelming the annotators, it is important to consider that the accuracies may be slightly influenced by the specific question sets sampled randomly. Future studies could expand the number of questions to provide a more comprehensive evaluation.

This study specifically focuses on ChatGPT and does not explore other LLMs like GPT-4, LLaMA \cite{touvron2023llama} and Google's Bard \cite{thoppilan2022lamda}. We choose ChatGPT in this study to achieve a good balance between popularity, availability, and cost. It would be interesting for future research to explore whether similar findings hold true for these models and to compare their performance against ChatGPT.

\section*{Acknowledgements}

This work is supported by the Natural Science Foundation of China (No. 62122077 and 62106251).

\bibliography{custom1}

\begin{thebibliography}{59}
\expandafter\ifx\csname natexlab\endcsname\relax\def\natexlab#1{#1}\fi

\bibitem[{Aggarwal et~al.(2021)Aggarwal, Mandowara, Agrawal, Khandelwal,
  Singla, and Garg}]{aggarwal2021explanations}
Shourya Aggarwal, Divyanshu Mandowara, Vishwajeet Agrawal, Dinesh Khandelwal,
  Parag Singla, and Dinesh Garg. 2021.
\newblock \href {https://doi.org/10.18653/v1/2021.acl-long.238} {{E}xplanations
  for {C}ommonsense{QA}: {N}ew {D}ataset and {M}odels}.
\newblock In \emph{Proceedings of the 59th Annual Meeting of the Association
  for Computational Linguistics and the 11th International Joint Conference on
  Natural Language Processing (Volume 1: Long Papers)}, pages 3050--3065,
  Online. Association for Computational Linguistics.

\bibitem[{Bang et~al.(2023)Bang, Cahyawijaya, Lee, Dai, Su, Wilie, Lovenia, Ji,
  Yu, Chung et~al.}]{bang2023multitask}
Yejin Bang, Samuel Cahyawijaya, Nayeon Lee, Wenliang Dai, Dan Su, Bryan Wilie,
  Holy Lovenia, Ziwei Ji, Tiezheng Yu, Willy Chung, et~al. 2023.
\newblock \href {https://arxiv.org/abs/2302.04023} {A multitask, multilingual,
  multimodal evaluation of chatgpt on reasoning, hallucination, and
  interactivity}.
\newblock \emph{ArXiv preprint}, abs/2302.04023.

\bibitem[{Bauer et~al.(2018)Bauer, Wang, and Bansal}]{bauer2018commonsense}
Lisa Bauer, Yicheng Wang, and Mohit Bansal. 2018.
\newblock \href {https://doi.org/10.18653/v1/D18-1454} {Commonsense for
  generative multi-hop question answering tasks}.
\newblock In \emph{Proceedings of the 2018 Conference on Empirical Methods in
  Natural Language Processing}, pages 4220--4230, Brussels, Belgium.
  Association for Computational Linguistics.

\bibitem[{Bender(2015)}]{bender2015establishing}
David Bender. 2015.
\newblock Establishing a human baseline for the winograd schema challenge.
\newblock In \emph{MAICS}, pages 39--45.

\bibitem[{Bhargava and Ng(2022)}]{bhargava2022commonsense}
Prajjwal Bhargava and Vincent Ng. 2022.
\newblock \href {https://ojs.aaai.org/index.php/AAAI/article/view/21496}
  {Commonsense knowledge reasoning and generation with pre-trained language
  models: {A} survey}.
\newblock In \emph{Thirty-Sixth {AAAI} Conference on Artificial Intelligence,
  {AAAI} 2022, Thirty-Fourth Conference on Innovative Applications of
  Artificial Intelligence, {IAAI} 2022, The Twelveth Symposium on Educational
  Advances in Artificial Intelligence, {EAAI} 2022 Virtual Event, February 22 -
  March 1, 2022}, pages 12317--12325. {AAAI} Press.

\bibitem[{Bisk et~al.(2020)Bisk, Zellers, LeBras, Gao, and Choi}]{bisk2020piqa}
Yonatan Bisk, Rowan Zellers, Ronan LeBras, Jianfeng Gao, and Yejin Choi. 2020.
\newblock \href {https://aaai.org/ojs/index.php/AAAI/article/view/6239}
  {{PIQA:} reasoning about physical commonsense in natural language}.
\newblock In \emph{The Thirty-Fourth {AAAI} Conference on Artificial
  Intelligence, {AAAI} 2020, The Thirty-Second Innovative Applications of
  Artificial Intelligence Conference, {IAAI} 2020, The Tenth {AAAI} Symposium
  on Educational Advances in Artificial Intelligence, {EAAI} 2020, New York,
  NY, USA, February 7-12, 2020}, pages 7432--7439. {AAAI} Press.

\bibitem[{Boratko et~al.(2020)Boratko, Li, O{'}Gorman, Das, Le, and
  McCallum}]{boratko2020protoqa}
Michael Boratko, Xiang Li, Tim O{'}Gorman, Rajarshi Das, Dan Le, and Andrew
  McCallum. 2020.
\newblock \href {https://doi.org/10.18653/v1/2020.emnlp-main.85} {{P}roto{QA}:
  A question answering dataset for prototypical common-sense reasoning}.
\newblock In \emph{Proceedings of the 2020 Conference on Empirical Methods in
  Natural Language Processing (EMNLP)}, pages 1122--1136, Online. Association
  for Computational Linguistics.

\bibitem[{Brachman and Levesque(2022)}]{brachman2022new}
Ronald~J. Brachman and Hector~J. Levesque. 2022.
\newblock \href {https://ojs.aaai.org/index.php/AAAI/article/view/21485}
  {Toward a new science of common sense}.
\newblock In \emph{Thirty-Sixth {AAAI} Conference on Artificial Intelligence,
  {AAAI} 2022, Thirty-Fourth Conference on Innovative Applications of
  Artificial Intelligence, {IAAI} 2022, The Twelveth Symposium on Educational
  Advances in Artificial Intelligence, {EAAI} 2022 Virtual Event, February 22 -
  March 1, 2022}, pages 12245--12249. {AAAI} Press.

\bibitem[{Brown et~al.(2020)Brown, Mann, Ryder, Subbiah, Kaplan, Dhariwal,
  Neelakantan, Shyam, Sastry, Askell, Agarwal, Herbert{-}Voss, Krueger,
  Henighan, Child, Ramesh, Ziegler, Wu, Winter, Hesse, Chen, Sigler, Litwin,
  Gray, Chess, Clark, Berner, McCandlish, Radford, Sutskever, and
  Amodei}]{brown2020language}
Tom~B. Brown, Benjamin Mann, Nick Ryder, Melanie Subbiah, Jared Kaplan,
  Prafulla Dhariwal, Arvind Neelakantan, Pranav Shyam, Girish Sastry, Amanda
  Askell, Sandhini Agarwal, Ariel Herbert{-}Voss, Gretchen Krueger, Tom
  Henighan, Rewon Child, Aditya Ramesh, Daniel~M. Ziegler, Jeffrey Wu, Clemens
  Winter, Christopher Hesse, Mark Chen, Eric Sigler, Mateusz Litwin, Scott
  Gray, Benjamin Chess, Jack Clark, Christopher Berner, Sam McCandlish, Alec
  Radford, Ilya Sutskever, and Dario Amodei. 2020.
\newblock \href
  {https://proceedings.neurips.cc/paper/2020/hash/1457c0d6bfcb4967418bfb8ac142f64a-Abstract.html}
  {Language models are few-shot learners}.
\newblock In \emph{Advances in Neural Information Processing Systems 33: Annual
  Conference on Neural Information Processing Systems 2020, NeurIPS 2020,
  December 6-12, 2020, virtual}.

\bibitem[{Chang and Bergen(2023)}]{chang2023language}
Tyler~A Chang and Benjamin~K Bergen. 2023.
\newblock \href {https://arxiv.org/abs/2303.11504} {Language model behavior: A
  comprehensive survey}.
\newblock \emph{ArXiv preprint}, abs/2303.11504.

\bibitem[{Chen et~al.(2023)Chen, Xu, Yan, Zhang, Huang, Si, and
  Zhang}]{chen2023distinguish}
Qianglong Chen, Guohai Xu, Ming Yan, Ji~Zhang, Fei Huang, Luo Si, and Yin
  Zhang. 2023.
\newblock \href {https://arxiv.org/abs/2305.08135} {Distinguish before answer:
  Generating contrastive explanation as knowledge for commonsense question
  answering}.
\newblock \emph{ArXiv preprint}, abs/2305.08135.

\bibitem[{Clark et~al.(2018)Clark, Cowhey, Etzioni, Khot, Sabharwal, Schoenick,
  and Tafjord}]{clark2018think}
Peter Clark, Isaac Cowhey, Oren Etzioni, Tushar Khot, Ashish Sabharwal, Carissa
  Schoenick, and Oyvind Tafjord. 2018.
\newblock \href {https://arxiv.org/abs/1803.05457} {Think you have solved
  question answering? try arc, the ai2 reasoning challenge}.
\newblock \emph{ArXiv preprint}, abs/1803.05457.

\bibitem[{Cohn and Hernandez-Orallo(2023)}]{cohn2023dialectical}
Anthony~G Cohn and Jose Hernandez-Orallo. 2023.
\newblock \href {https://arxiv.org/abs/2304.11164} {Dialectical language model
  evaluation: An initial appraisal of the commonsense spatial reasoning
  abilities of llms}.
\newblock \emph{ArXiv preprint}, abs/2304.11164.

\bibitem[{Cui et~al.(2021)Cui, Cheng, Wu, and Zhang}]{cui2021commonsense}
Leyang Cui, Sijie Cheng, Yu~Wu, and Yue Zhang. 2021.
\newblock \href {https://doi.org/10.18653/v1/2021.findings-acl.61} {On
  commonsense cues in {BERT} for solving commonsense tasks}.
\newblock In \emph{Findings of the Association for Computational Linguistics:
  ACL-IJCNLP 2021}, pages 683--693, Online. Association for Computational
  Linguistics.

\bibitem[{Cui and Chen(2023)}]{cui2023free}
Wanyun Cui and Xingran Chen. 2023.
\newblock \href {https://arxiv.org/abs/2305.15516} {Free lunch for efficient
  textual commonsense integration in language models}.
\newblock \emph{ArXiv preprint}, abs/2305.15516.

\bibitem[{Davis(2023)}]{davis2023benchmarks}
Ernest Davis. 2023.
\newblock \href {https://arxiv.org/abs/2302.04752} {Benchmarks for automated
  commonsense reasoning: A survey}.
\newblock \emph{ArXiv preprint}, abs/2302.04752.

\bibitem[{Davison et~al.(2019)Davison, Feldman, and
  Rush}]{davison2019commonsense}
Joe Davison, Joshua Feldman, and Alexander Rush. 2019.
\newblock \href {https://doi.org/10.18653/v1/D19-1109} {Commonsense knowledge
  mining from pretrained models}.
\newblock In \emph{Proceedings of the 2019 Conference on Empirical Methods in
  Natural Language Processing and the 9th International Joint Conference on
  Natural Language Processing (EMNLP-IJCNLP)}, pages 1173--1178, Hong Kong,
  China. Association for Computational Linguistics.

\bibitem[{Ellis(2008)}]{ellis2008implicit}
Nick~C Ellis. 2008.
\newblock Implicit and explicit knowledge about language.
\newblock \emph{Encyclopedia of language and education}, 6:1--13.

\bibitem[{Gordon and Van~Durme(2013)}]{gordon2013reporting}
Jonathan Gordon and Benjamin Van~Durme. 2013.
\newblock Reporting bias and knowledge acquisition.
\newblock In \emph{Proceedings of the 2013 workshop on Automated knowledge base
  construction}, pages 25--30.

\bibitem[{Gu et~al.(2022)Gu, Mishra, and Clark}]{gu2022language}
Yuling Gu, Bhavana~Dalvi Mishra, and Peter Clark. 2022.
\newblock \href {https://arxiv.org/abs/2212.10029} {Do language models have
  coherent mental models of everyday things?}
\newblock \emph{ArXiv preprint}, abs/2212.10029.

\bibitem[{He et~al.(2023)He, Guti{\'e}rrez-Basulto, Pan et~al.}]{he2023buca}
Jie He, V{\'\i}ctor Guti{\'e}rrez-Basulto, Jeff~Z Pan, et~al. 2023.
\newblock \href {https://arxiv.org/abs/2305.15932} {Buca: A binary
  classification approach to unsupervised commonsense question answering}.
\newblock \emph{ArXiv preprint}, abs/2305.15932.

\bibitem[{Jiang et~al.(2021)Jiang, Bosselut, Bhagavatula, and
  Choi}]{jiang2021m}
Liwei Jiang, Antoine Bosselut, Chandra Bhagavatula, and Yejin Choi. 2021.
\newblock \href {https://doi.org/10.18653/v1/2021.naacl-main.346} {{``}{I}{'}m
  not mad{''}: Commonsense implications of negation and contradiction}.
\newblock In \emph{Proceedings of the 2021 Conference of the North American
  Chapter of the Association for Computational Linguistics: Human Language
  Technologies}, pages 4380--4397, Online. Association for Computational
  Linguistics.

\bibitem[{Khot et~al.(2020)Khot, Clark, Guerquin, Jansen, and
  Sabharwal}]{khot2020qasc}
Tushar Khot, Peter Clark, Michal Guerquin, Peter Jansen, and Ashish Sabharwal.
  2020.
\newblock \href {https://aaai.org/ojs/index.php/AAAI/article/view/6319}
  {{QASC:} {A} dataset for question answering via sentence composition}.
\newblock In \emph{The Thirty-Fourth {AAAI} Conference on Artificial
  Intelligence, {AAAI} 2020, The Thirty-Second Innovative Applications of
  Artificial Intelligence Conference, {IAAI} 2020, The Tenth {AAAI} Symposium
  on Educational Advances in Artificial Intelligence, {EAAI} 2020, New York,
  NY, USA, February 7-12, 2020}, pages 8082--8090. {AAAI} Press.

\bibitem[{Klein and Nabi(2021)}]{klein2021towards}
Tassilo Klein and Moin Nabi. 2021.
\newblock \href {https://doi.org/10.18653/v1/2021.emnlp-main.688} {Towards
  zero-shot commonsense reasoning with self-supervised refinement of language
  models}.
\newblock In \emph{Proceedings of the 2021 Conference on Empirical Methods in
  Natural Language Processing}, pages 8737--8743, Online and Punta Cana,
  Dominican Republic. Association for Computational Linguistics.

\bibitem[{Kondo et~al.(2023)Kondo, Sugawara, and Aizawa}]{kondo2023probing}
Kazushi Kondo, Saku Sugawara, and Akiko Aizawa. 2023.
\newblock \href {https://arxiv.org/abs/2306.02258} {Commonsense knowledge
  transfer for pre-trained language models}.
\newblock \emph{ArXiv preprint}, abs/2306.02258.

\bibitem[{Laskar et~al.(2023)Laskar, Bari, Rahman, Bhuiyan, Joty, and
  Huang}]{laskar2023systematic}
Md~Tahmid~Rahman Laskar, M~Saiful Bari, Mizanur Rahman, Md~Amran~Hossen
  Bhuiyan, Shafiq Joty, and Jimmy~Xiangji Huang. 2023.
\newblock \href {https://arxiv.org/abs/2305.18486} {A systematic study and
  comprehensive evaluation of chatgpt on benchmark datasets}.
\newblock \emph{ArXiv preprint}, abs/2305.18486.

\bibitem[{Levesque et~al.(2012)Levesque, Davis, and
  Morgenstern}]{levesque2012winograd}
Hector~J Levesque, Ernest Davis, and Leora Morgenstern. 2012.
\newblock The winograd schema challenge.
\newblock In \emph{Proceedings of the Thirteenth International Conference on
  Principles of Knowledge Representation and Reasoning}, pages 552--561.

\bibitem[{Li et~al.(2021)Li, Kuncoro, d'Autume, Blunsom, and
  Nematzadeh}]{li2021language}
Xiang~Lorraine Li, Adhiguna Kuncoro, Cyprien de~Masson d'Autume, Phil Blunsom,
  and Aida Nematzadeh. 2021.
\newblock \href {https://arxiv.org/abs/2111.00607} {Do language models learn
  commonsense knowledge?}
\newblock \emph{ArXiv preprint}, abs/2111.00607.

\bibitem[{Li et~al.(2022)Li, Kuncoro, Hoffmann, de~Masson~d{'}Autume, Blunsom,
  and Nematzadeh}]{li2022systematic}
Xiang~Lorraine Li, Adhiguna Kuncoro, Jordan Hoffmann, Cyprien
  de~Masson~d{'}Autume, Phil Blunsom, and Aida Nematzadeh. 2022.
\newblock \href {https://aclanthology.org/2022.emnlp-main.812} {A systematic
  investigation of commonsense knowledge in large language models}.
\newblock In \emph{Proceedings of the 2022 Conference on Empirical Methods in
  Natural Language Processing}, pages 11838--11855, Abu Dhabi, United Arab
  Emirates. Association for Computational Linguistics.

\bibitem[{Lin et~al.(2020)Lin, Lee, Khanna, and Ren}]{lin2020birds}
Bill~Yuchen Lin, Seyeon Lee, Rahul Khanna, and Xiang Ren. 2020.
\newblock \href {https://doi.org/10.18653/v1/2020.emnlp-main.557} {{B}irds have
  four legs?! {N}umer{S}ense: {P}robing {N}umerical {C}ommonsense {K}nowledge
  of {P}re-{T}rained {L}anguage {M}odels}.
\newblock In \emph{Proceedings of the 2020 Conference on Empirical Methods in
  Natural Language Processing (EMNLP)}, pages 6862--6868, Online. Association
  for Computational Linguistics.

\bibitem[{Liu and Singh(2004)}]{liu2004conceptnet}
Hugo Liu and Push Singh. 2004.
\newblock Conceptnet—a practical commonsense reasoning tool-kit.
\newblock \emph{BT technology journal}, 22:211--226.

\bibitem[{Liu et~al.(2022)Liu, Liu, Lu, Welleck, West, Le~Bras, Choi, and
  Hajishirzi}]{liu2022generated}
Jiacheng Liu, Alisa Liu, Ximing Lu, Sean Welleck, Peter West, Ronan Le~Bras,
  Yejin Choi, and Hannaneh Hajishirzi. 2022.
\newblock \href {https://doi.org/10.18653/v1/2022.acl-long.225} {Generated
  knowledge prompting for commonsense reasoning}.
\newblock In \emph{Proceedings of the 60th Annual Meeting of the Association
  for Computational Linguistics (Volume 1: Long Papers)}, pages 3154--3169,
  Dublin, Ireland. Association for Computational Linguistics.

\bibitem[{Liu et~al.(2021)Liu, Wan, He, Peng, and Yu}]{liu2021kg}
Ye~Liu, Yao Wan, Lifang He, Hao Peng, and Philip~S. Yu. 2021.
\newblock \href {https://ojs.aaai.org/index.php/AAAI/article/view/16796}
  {{KG-BART:} knowledge graph-augmented {BART} for generative commonsense
  reasoning}.
\newblock In \emph{Thirty-Fifth {AAAI} Conference on Artificial Intelligence,
  {AAAI} 2021, Thirty-Third Conference on Innovative Applications of Artificial
  Intelligence, {IAAI} 2021, The Eleventh Symposium on Educational Advances in
  Artificial Intelligence, {EAAI} 2021, Virtual Event, February 2-9, 2021},
  pages 6418--6425. {AAAI} Press.

\bibitem[{Liu et~al.(2020)Liu, Yang, You, Fan, and Yu}]{liu2020commonsense}
Ye~Liu, Tao Yang, Zeyu You, Wei Fan, and Philip~S. Yu. 2020.
\newblock \href {https://aclanthology.org/2020.sigdial-1.9} {Commonsense
  evidence generation and injection in reading comprehension}.
\newblock In \emph{Proceedings of the 21th Annual Meeting of the Special
  Interest Group on Discourse and Dialogue}, pages 61--73, 1st virtual meeting.
  Association for Computational Linguistics.

\bibitem[{Ma et~al.(2021)Ma, Ilievski, Francis, Ozaki, Nyberg, and
  Oltramari}]{ma2021exploring}
Kaixin Ma, Filip Ilievski, Jonathan Francis, Satoru Ozaki, Eric Nyberg, and
  Alessandro Oltramari. 2021.
\newblock \href {https://doi.org/10.18653/v1/2021.emnlp-main.445} {Exploring
  strategies for generalizable commonsense reasoning with pre-trained models}.
\newblock In \emph{Proceedings of the 2021 Conference on Empirical Methods in
  Natural Language Processing}, pages 5474--5483, Online and Punta Cana,
  Dominican Republic. Association for Computational Linguistics.

\bibitem[{Mihaylov et~al.(2018)Mihaylov, Clark, Khot, and
  Sabharwal}]{mihaylov2018can}
Todor Mihaylov, Peter Clark, Tushar Khot, and Ashish Sabharwal. 2018.
\newblock \href {https://doi.org/10.18653/v1/D18-1260} {Can a suit of armor
  conduct electricity? a new dataset for open book question answering}.
\newblock In \emph{Proceedings of the 2018 Conference on Empirical Methods in
  Natural Language Processing}, pages 2381--2391, Brussels, Belgium.
  Association for Computational Linguistics.

\bibitem[{Niu et~al.(2021)Niu, Huang, Liang, Chen, Zhu, and
  Huang}]{niu2021semantic}
Yilin Niu, Fei Huang, Jiaming Liang, Wenkai Chen, Xiaoyan Zhu, and Minlie
  Huang. 2021.
\newblock \href {https://doi.org/10.18653/v1/2021.acl-long.237} {A
  semantic-based method for unsupervised commonsense question answering}.
\newblock In \emph{Proceedings of the 59th Annual Meeting of the Association
  for Computational Linguistics and the 11th International Joint Conference on
  Natural Language Processing (Volume 1: Long Papers)}, pages 3037--3049,
  Online. Association for Computational Linguistics.

\bibitem[{Ouyang et~al.(2022)Ouyang, Wu, Jiang, Almeida, Wainwright, Mishkin,
  Zhang, Agarwal, Slama, Ray et~al.}]{ouyang2022training}
Long Ouyang, Jeff Wu, Xu~Jiang, Diogo Almeida, Carroll~L Wainwright, Pamela
  Mishkin, Chong Zhang, Sandhini Agarwal, Katarina Slama, Alex Ray, et~al.
  2022.
\newblock \href {https://arxiv.org/abs/2203.02155} {Training language models to
  follow instructions with human feedback}.
\newblock \emph{ArXiv preprint}, abs/2203.02155.

\bibitem[{Porada et~al.(2022)Porada, Sordoni, and Cheung}]{porada2021does}
Ian Porada, Alessandro Sordoni, and Jackie Cheung. 2022.
\newblock \href {https://doi.org/10.18653/v1/2022.naacl-main.337} {Does
  pre-training induce systematic inference? how masked language models acquire
  commonsense knowledge}.
\newblock In \emph{Proceedings of the 2022 Conference of the North American
  Chapter of the Association for Computational Linguistics: Human Language
  Technologies}, pages 4550--4557, Seattle, United States. Association for
  Computational Linguistics.

\bibitem[{Rajani et~al.(2019)Rajani, McCann, Xiong, and
  Socher}]{rajani2019explain}
Nazneen~Fatema Rajani, Bryan McCann, Caiming Xiong, and Richard Socher. 2019.
\newblock \href {https://doi.org/10.18653/v1/P19-1487} {Explain yourself!
  leveraging language models for commonsense reasoning}.
\newblock In \emph{Proceedings of the 57th Annual Meeting of the Association
  for Computational Linguistics}, pages 4932--4942, Florence, Italy.
  Association for Computational Linguistics.

\bibitem[{Sap et~al.(2019)Sap, Rashkin, Chen, Le~Bras, and
  Choi}]{sap2019social}
Maarten Sap, Hannah Rashkin, Derek Chen, Ronan Le~Bras, and Yejin Choi. 2019.
\newblock \href {https://doi.org/10.18653/v1/D19-1454} {Social {IQ}a:
  Commonsense reasoning about social interactions}.
\newblock In \emph{Proceedings of the 2019 Conference on Empirical Methods in
  Natural Language Processing and the 9th International Joint Conference on
  Natural Language Processing (EMNLP-IJCNLP)}, pages 4463--4473, Hong Kong,
  China. Association for Computational Linguistics.

\bibitem[{Shwartz and Choi(2020)}]{shwartz2020neural}
Vered Shwartz and Yejin Choi. 2020.
\newblock \href {https://doi.org/10.18653/v1/2020.coling-main.605} {Do neural
  language models overcome reporting bias?}
\newblock In \emph{Proceedings of the 28th International Conference on
  Computational Linguistics}, pages 6863--6870, Barcelona, Spain (Online).
  International Committee on Computational Linguistics.

\bibitem[{Shwartz et~al.(2020)Shwartz, West, Le~Bras, Bhagavatula, and
  Choi}]{shwartz2020unsupervised}
Vered Shwartz, Peter West, Ronan Le~Bras, Chandra Bhagavatula, and Yejin Choi.
  2020.
\newblock \href {https://doi.org/10.18653/v1/2020.emnlp-main.373} {Unsupervised
  commonsense question answering with self-talk}.
\newblock In \emph{Proceedings of the 2020 Conference on Empirical Methods in
  Natural Language Processing (EMNLP)}, pages 4615--4629, Online. Association
  for Computational Linguistics.

\bibitem[{Sun et~al.(2022)Sun, Zhang, Qi, and Shi}]{sun2022tsgp}
Yueqing Sun, Yu~Zhang, Le~Qi, and Qi~Shi. 2022.
\newblock \href {https://aclanthology.org/2022.findings-emnlp.68} {{TSGP}:
  Two-stage generative prompting for unsupervised commonsense question
  answering}.
\newblock In \emph{Findings of the Association for Computational Linguistics:
  EMNLP 2022}, pages 968--980, Abu Dhabi, United Arab Emirates. Association for
  Computational Linguistics.

\bibitem[{Talmor et~al.(2019)Talmor, Herzig, Lourie, and
  Berant}]{talmor2019commonsenseqa}
Alon Talmor, Jonathan Herzig, Nicholas Lourie, and Jonathan Berant. 2019.
\newblock \href {https://doi.org/10.18653/v1/N19-1421} {{C}ommonsense{QA}: A
  question answering challenge targeting commonsense knowledge}.
\newblock In \emph{Proceedings of the 2019 Conference of the North {A}merican
  Chapter of the Association for Computational Linguistics: Human Language
  Technologies, Volume 1 (Long and Short Papers)}, pages 4149--4158,
  Minneapolis, Minnesota. Association for Computational Linguistics.

\bibitem[{Tamborrino et~al.(2020)Tamborrino, Pellican{\`o}, Pannier, Voitot,
  and Naudin}]{tamborrino2020pre}
Alexandre Tamborrino, Nicola Pellican{\`o}, Baptiste Pannier, Pascal Voitot,
  and Louise Naudin. 2020.
\newblock \href {https://doi.org/10.18653/v1/2020.acl-main.357} {Pre-training
  is (almost) all you need: An application to commonsense reasoning}.
\newblock In \emph{Proceedings of the 58th Annual Meeting of the Association
  for Computational Linguistics}, pages 3878--3887, Online. Association for
  Computational Linguistics.

\bibitem[{Tang et~al.(2023)Tang, Zheng, Li, Meng, Zhu, Liang, and
  Zhang}]{tang2023large}
Xiaojuan Tang, Zilong Zheng, Jiaqi Li, Fanxu Meng, Song-Chun Zhu, Yitao Liang,
  and Muhan Zhang. 2023.
\newblock \href {https://arxiv.org/abs/2305.14825} {Large language models are
  in-context semantic reasoners rather than symbolic reasoners}.
\newblock \emph{ArXiv preprint}, abs/2305.14825.

\bibitem[{Thoppilan et~al.(2022)Thoppilan, De~Freitas, Hall, Shazeer,
  Kulshreshtha, Cheng, Jin, Bos, Baker, Du et~al.}]{thoppilan2022lamda}
Romal Thoppilan, Daniel De~Freitas, Jamie Hall, Noam Shazeer, Apoorv
  Kulshreshtha, Heng-Tze Cheng, Alicia Jin, Taylor Bos, Leslie Baker, Yu~Du,
  et~al. 2022.
\newblock \href {https://arxiv.org/abs/2201.08239} {Lamda: Language models for
  dialog applications}.
\newblock \emph{ArXiv preprint}, abs/2201.08239.

\bibitem[{Touvron et~al.(2023)Touvron, Lavril, Izacard, Martinet, Lachaux,
  Lacroix, Rozi{\`e}re, Goyal, Hambro, Azhar, Rodriguez, Joulin, Grave, and
  Lample}]{touvron2023llama}
Hugo Touvron, Thibaut Lavril, Gautier Izacard, Xavier Martinet, Marie-Anne
  Lachaux, Timoth{\'e}e Lacroix, Baptiste Rozi{\`e}re, Naman Goyal, Eric
  Hambro, Faisal Azhar, Aurelien Rodriguez, Armand Joulin, Edouard Grave, and
  Guillaume Lample. 2023.
\newblock \href {https://arxiv.org/abs/2302.13971} {Llama: Open and efficient
  foundation language models}.
\newblock \emph{ArXiv preprint}, abs/2302.13971.

\bibitem[{Wang et~al.(2021)Wang, Ilievski, Chen, and Ren}]{wang2021language}
Peifeng Wang, Filip Ilievski, Muhao Chen, and Xiang Ren. 2021.
\newblock \href {https://doi.org/10.18653/v1/2021.findings-acl.322} {Do
  language models perform generalizable commonsense inference?}
\newblock In \emph{Findings of the Association for Computational Linguistics:
  ACL-IJCNLP 2021}, pages 3681--3688, Online. Association for Computational
  Linguistics.

\bibitem[{Wang et~al.(2020)Wang, Peng, Ilievski, Szekely, and
  Ren}]{wang2020connecting}
Peifeng Wang, Nanyun Peng, Filip Ilievski, Pedro Szekely, and Xiang Ren. 2020.
\newblock \href {https://doi.org/10.18653/v1/2020.findings-emnlp.369}
  {Connecting the dots: A knowledgeable path generator for commonsense question
  answering}.
\newblock In \emph{Findings of the Association for Computational Linguistics:
  EMNLP 2020}, pages 4129--4140, Online. Association for Computational
  Linguistics.

\bibitem[{Wei et~al.(2022)Wei, Wang, Schuurmans, Bosma, Chi, Le, and
  Zhou}]{wei2022chain}
Jason Wei, Xuezhi Wang, Dale Schuurmans, Maarten Bosma, Ed~Chi, Quoc Le, and
  Denny Zhou. 2022.
\newblock \href {https://arxiv.org/abs/2201.11903} {Chain of thought prompting
  elicits reasoning in large language models}.
\newblock \emph{ArXiv preprint}, abs/2201.11903.

\bibitem[{West et~al.(2022)West, Bhagavatula, Hessel, Hwang, Jiang, Le~Bras,
  Lu, Welleck, and Choi}]{west2022symbolic}
Peter West, Chandra Bhagavatula, Jack Hessel, Jena Hwang, Liwei Jiang, Ronan
  Le~Bras, Ximing Lu, Sean Welleck, and Yejin Choi. 2022.
\newblock \href {https://doi.org/10.18653/v1/2022.naacl-main.341} {Symbolic
  knowledge distillation: from general language models to commonsense models}.
\newblock In \emph{Proceedings of the 2022 Conference of the North American
  Chapter of the Association for Computational Linguistics: Human Language
  Technologies}, pages 4602--4625, Seattle, United States. Association for
  Computational Linguistics.

\bibitem[{Zellers et~al.(2019)Zellers, Holtzman, Bisk, Farhadi, and
  Choi}]{zellers2019hellaswag}
Rowan Zellers, Ari Holtzman, Yonatan Bisk, Ali Farhadi, and Yejin Choi. 2019.
\newblock \href {https://doi.org/10.18653/v1/P19-1472} {{H}ella{S}wag: Can a
  machine really finish your sentence?}
\newblock In \emph{Proceedings of the 57th Annual Meeting of the Association
  for Computational Linguistics}, pages 4791--4800, Florence, Italy.
  Association for Computational Linguistics.

\bibitem[{Zhang et~al.(2022)Zhang, Li, Wu, Su, and Sun}]{zhang2022alleviating}
Yi~Zhang, Lei Li, Yunfang Wu, Qi~Su, and Xu~Sun. 2022.
\newblock Alleviating the knowledge-language inconsistency: A study for deep
  commonsense knowledge.
\newblock \emph{IEEE/ACM Transactions on Audio, Speech, and Language
  Processing}, 30:594--604.

\bibitem[{Zhao et~al.(2023)Zhao, Lee, and Hsu}]{zhao2023large}
Zirui Zhao, Wee~Sun Lee, and David Hsu. 2023.
\newblock \href {https://arxiv.org/abs/2305.14078} {Large language models as
  commonsense knowledge for large-scale task planning}.
\newblock \emph{ArXiv preprint}, abs/2305.14078.

\bibitem[{Zhou et~al.(2019)Zhou, Khashabi, Ning, and Roth}]{zhou2019going}
Ben Zhou, Daniel Khashabi, Qiang Ning, and Dan Roth. 2019.
\newblock \href {https://doi.org/10.18653/v1/D19-1332} {{``}going on a
  vacation{''} takes longer than {``}going for a walk{''}: A study of temporal
  commonsense understanding}.
\newblock In \emph{Proceedings of the 2019 Conference on Empirical Methods in
  Natural Language Processing and the 9th International Joint Conference on
  Natural Language Processing (EMNLP-IJCNLP)}, pages 3363--3369, Hong Kong,
  China. Association for Computational Linguistics.

\bibitem[{Zhou et~al.(2023)Zhou, Bras, and Choi}]{zhou2023commonsense}
Wangchunshu Zhou, Ronan~Le Bras, and Yejin Choi. 2023.
\newblock \href {https://arxiv.org/abs/2306.02388} {Commonsense knowledge
  transfer for pre-trained language models}.
\newblock \emph{ArXiv preprint}, abs/2306.02388.

\bibitem[{Zhou et~al.(2020)Zhou, Zhang, Cui, and Huang}]{zhou2020evaluating}
Xuhui Zhou, Yue Zhang, Leyang Cui, and Dandan Huang. 2020.
\newblock \href {https://aaai.org/ojs/index.php/AAAI/article/view/6523}
  {Evaluating commonsense in pre-trained language models}.
\newblock In \emph{The Thirty-Fourth {AAAI} Conference on Artificial
  Intelligence, {AAAI} 2020, The Thirty-Second Innovative Applications of
  Artificial Intelligence Conference, {IAAI} 2020, The Tenth {AAAI} Symposium
  on Educational Advances in Artificial Intelligence, {EAAI} 2020, New York,
  NY, USA, February 7-12, 2020}, pages 9733--9740. {AAAI} Press.

\end{thebibliography}
\bibliographystyle{acl_natbib}

\end{document}